%% file: main.tex
\documentclass[10pt,twocolumn,letterpaper]{article}

\usepackage[pagenumbers]{iccv} 

\input{preamble}

\definecolor{myGreen}{rgb}{0, .6, .0}
\definecolor{myPink}{rgb}{0.87, 0.45, 0.45}

\definecolor{iccvblue}{rgb}{0.21,0.49,0.74}
\usepackage[pagebackref,breaklinks,colorlinks,allcolors=iccvblue]{hyperref}

\def\paperID{4354} 
\def\confName{ICCV}
\def\confYear{2025}

\title{Inpaint4Drag:  Repurposing Inpainting Models for Drag-Based Image Editing via Bidirectional Warping}

\author{Jingyi Lu \quad Kai Han$^{\dagger}$\\
Visual AI Lab, The University of Hong Kong\\
{\tt\small lujingyi@connect.hku.hk, kaihanx@hku.hk}
}

\begin{document}
\maketitle
\footnotetext[2]{Corresponding author.}

\addtocontents{toc}{\protect\setcounter{tocdepth}{-2}}

\input{sec/0_abstract}    
\input{sec/1_intro}
\input{sec/2_related_work}
\input{sec/3_method}

\input{sec/4_experiments}
\input{sec/5_conclusion}

{
    \small
    \bibliographystyle{ieeenat_fullname}
    \bibliography{main}
}

\clearpage
\onecolumn

\renewcommand{\thesection}{S\arabic{section}}
\setcounter{section}{0}

\setcounter{figure}{0}
\renewcommand{\thefigure}{S\arabic{figure}}

\setcounter{table}{0}
\renewcommand{\thetable}{S\arabic{table}}

\addtocontents{toc}{\protect\setcounter{tocdepth}{1}}

\begin{center}
{\Large \textbf{Inpaint4Drag:  Repurposing Inpainting Models for Drag-Based Image Editing via Bidirectional Warping}}\\
\vspace{0.5cm}
{\large \textit{-- Supplementary Material --}}\\
\vspace{0.5cm}
Jingyi Lu \quad Kai Han$^{\dagger}$\\
Visual AI Lab, The University of Hong Kong\\
{\tt\small lujingyi@connect.hku.hk, kaihanx@hku.hk}
\end{center}

\vspace{1cm}

{
\makeatletter
\renewcommand{\l@section}{\@dottedtocline{1}{1.5em}{2.0em}}
\makeatother
  \tableofcontents
}

\clearpage

\section{Supplementary Videos}
Two supplementary videos are available on our project page \url{https://visual-ai.github.io/inpaint4drag}: one demonstrating our bidirectional warping algorithm with visualizations, and another showcasing the real-time user interface interaction.

\section{Integration with More Inpainting Methods}
\label{sec:supp_inpaints}
\begin{figure}[ht]
    \centering
    \includegraphics[width=0.7\linewidth]{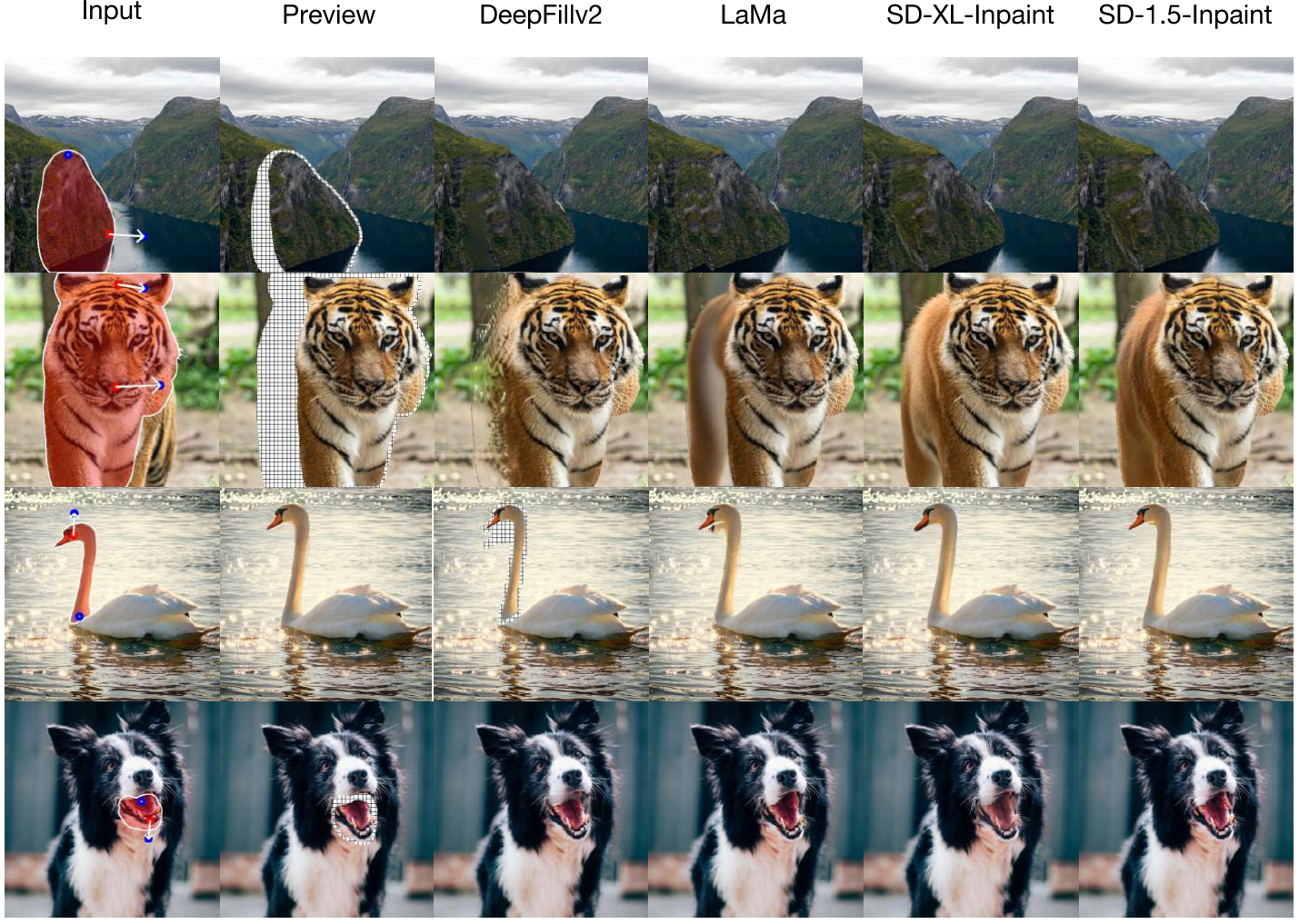}
    \caption{Qualitative comparison of different inpainting methods. The figure illustrates how various inpainting approaches affect drag editing results, highlighting differences in visual fidelity, artifact handling, and preservation of semantic content across traditional and generative model-based techniques.}
    \label{fig:inpaints}
\end{figure}

We integrate our framework with diverse image inpainting approaches, from early methods like LaMa \cite{suvorov2022resolution} and DeepFillv2 \cite{yu2019free} to recent generative model-based techniques \cite{rombach2022high}. While quantitative metrics in \cref{tab: inpaint} show comparable drag editing performance across methods, qualitative differences emerge in \cref{fig:inpaints}. Early approaches offer computational efficiency, whereas generative methods sometimes produce more realistic results--a quality distinction not fully captured by existing metrics. Our framework provides users flexibility to select the inpainting method best suited to their specific requirements, balancing computational resources and visual fidelity.

\begin{table}[h]
\centering
\begin{tabular}{@{\hspace{1pt}}l@{\hspace{4pt}}c@{\hspace{4pt}}c@{\hspace{4pt}}c@{\hspace{2pt}}c@{\hspace{4pt}}c@{\hspace{2pt}}c@{\hspace{1pt}}}
\toprule
Method & & & \multicolumn{2}{c}{DragBench-S} & \multicolumn{2}{c}{DragBench-D} \\
& Mem(GB)$\downarrow$ & Time(s)$\downarrow$ & MD$\downarrow$ & LPIPS$\downarrow$ & MD$\downarrow$ & LPIPS$\downarrow$ \\
\midrule
DeepFillv2 \cite{yu2019free} &\textbf{ 0.8 }& \textbf{0.05} & \textbf{3.2} & 13.7 & \textbf{3.7} & 9.0 \\
LaMa \cite{suvorov2022resolution} & 1.1 & 0.07 & 3.4 & 13.6 & 3.7 & 9.0 \\
SD-XL-Inpaint \cite{rombach2022high} & 8.1 & 1.3 & \textbf{3.2} & 12.5 & 3.8 & \textbf{8.8}\\
SD-1.5-Inpaint \cite{rombach2022high} & 2.7 & 0.3 & 3.6 & \textbf{11.4} & 3.9 & 9.1 \\
\bottomrule
\end{tabular}
\caption{Comparison of different inpainting methods. MD and LPIPS values are scaled by 100. Time and GPU memory are measured at 512$\times$512 resolution.}
\label{tab: inpaint}
\end{table}

\clearpage

\section{Multi-round Interactive Editing}
Our system enables fluid multi-round interactions, allowing users to execute sequential edits with minimal delay. In \cref{fig:chess_example}, we demonstrate this capability through a chess sequence where pieces are repositioned in rapid succession to create a checkmate scenario--highlighting our system's responsiveness to iterative user inputs.

\input{figs/chess}

\clearpage

\section{Discussion of Input Ambiguity}
\begin{figure}[h]
    \centering
    \includegraphics[width=0.98\linewidth]{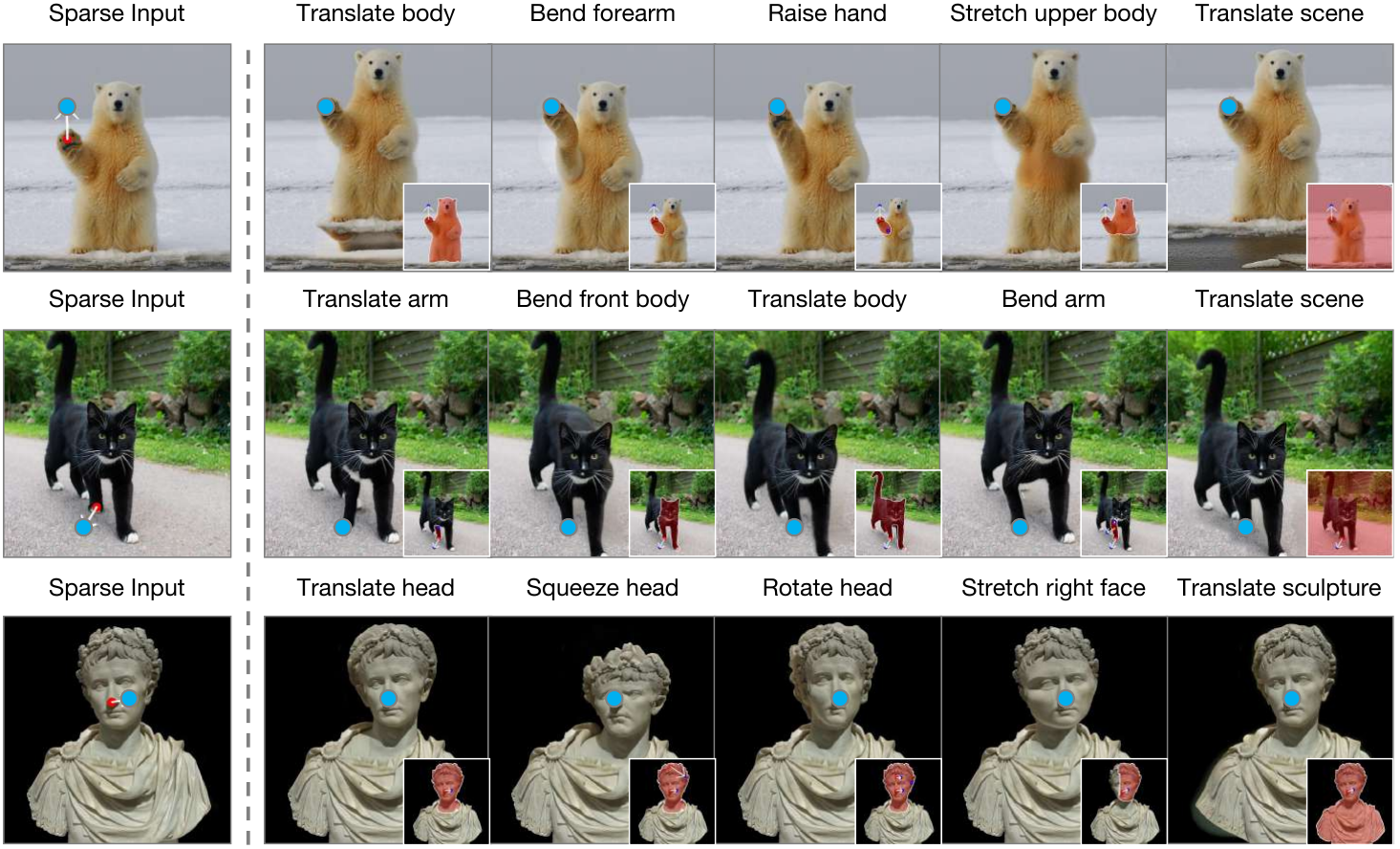}
    \caption{Precise control over ambiguous drag operations. \textbf{Left}: Ambiguous sparse input from previous methods can represent at least five different user intentions. \textbf{Right}: Through our explicit deformation-based control interface (bottom-right insets), we precisely implement each distinct user intention, effectively eliminating ambiguity while maintaining intuitive interaction.}
    \label{fig:ambiguity}
\end{figure}

Previous drag editing methods \cite{pan2023drag, shi2024instadrag, zhao2024fastdrag} typically use sparse control points to guide deformation, with optional masking to restrict editable regions. However, this sparse input format (shown on the left of each row in \cref{fig:ambiguity}) introduces fundamental ambiguity in deformation interpretation. Through our explicit region-based control (visualized in bottom-right insets), we demonstrate how a single ambiguous drag input can be precisely controlled to achieve five distinct editing intentions - from local manipulation to global translation. For instance, the same drag operation on a polar bear can be accurately interpreted as body translation, forearm bending, hand raising, upper body stretching, or scene translation. We address this ambiguity by requesting users to specify deformable regions through masking, treating each region as an elastic material where movement smoothly propagates from control points throughout the connected area.

\clearpage
\section{More Qualitative Results}
\begin{figure}[h]
    \centering
    \includegraphics[width=.9\linewidth]{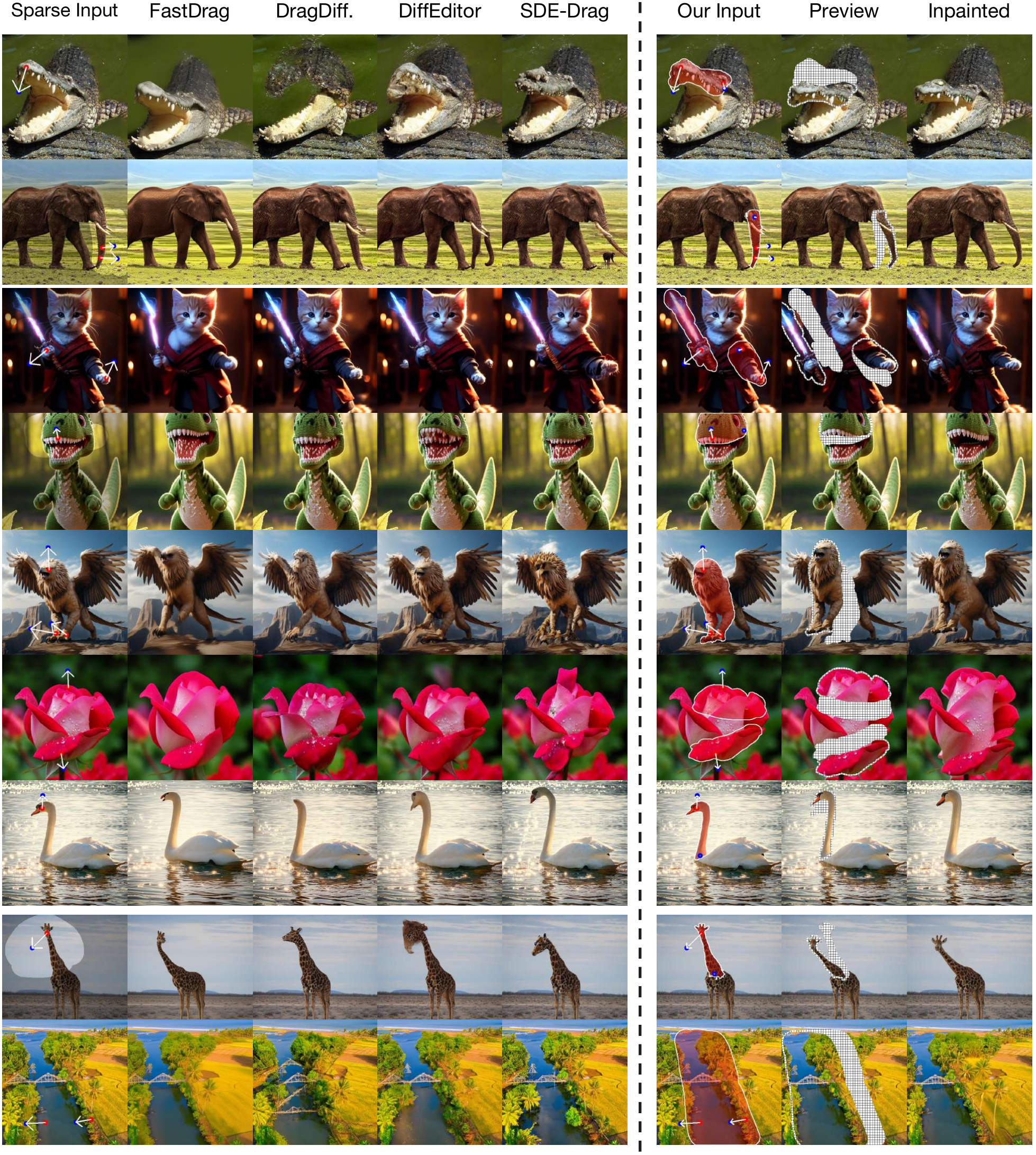}
    \caption{Qualitative comparison of Inpaint4Drag with state-of-the-art methods: wildlife, artworks, flowers, birds, and landscapes.}
    \label{fig:qualitative_supp_1}
\end{figure}

\begin{figure}[h]
    \centering
    \includegraphics[width=\linewidth]{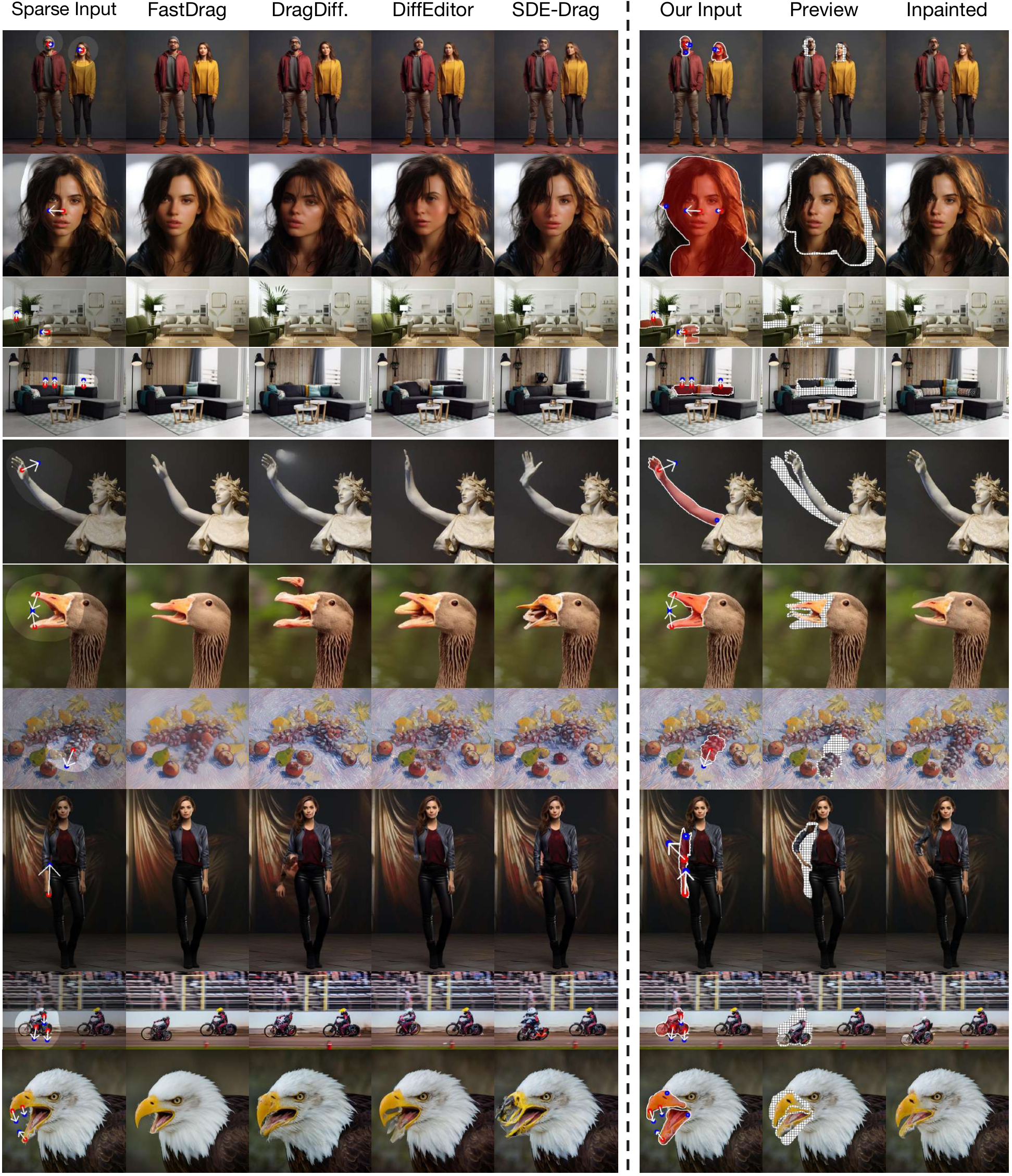}
    \caption{Qualitative comparison of Inpaint4Drag with state-of-the-art methods: portraits, interiors, statues, wildlife, still life, and sports.}
    \label{fig:qualitative_supp_2}
\end{figure}

\begin{figure}[h]
    \centering
    \includegraphics[width=\linewidth]{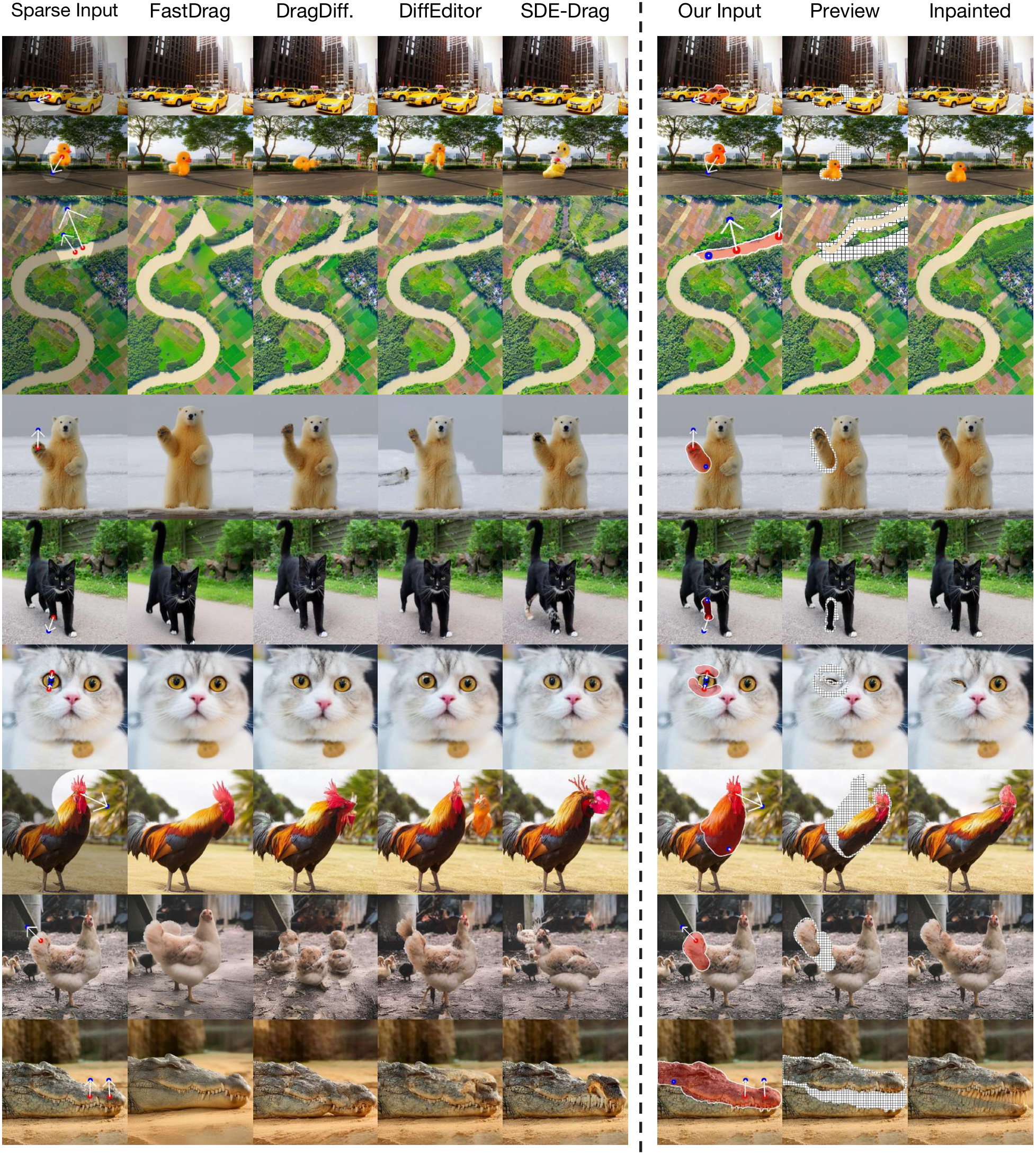}
    \caption{Qualitative comparison of Inpaint4Drag with state-of-the-art methods: urban scenes, landscapes, animals, pets, and reptiles.}
    \label{fig:qualitative_supp_3}
\end{figure}

We present extensive qualitative results in \cref{fig:qualitative_supp_1,fig:qualitative_supp_2,fig:qualitative_supp_3}. Our method allows users to specify handle points (\textcolor{red}{red}) and target points (\textcolor{blue}{blue}) with arrows defining deformation regions (highlighted in \textcolor{red}{red}). By applying elastic material principles directly in pixel space, we achieve superior performance across diverse editing scenarios. The results demonstrate our method's effectiveness in facial edits, large-scale deformations, and precise local manipulations while maintaining geometric stability. This advantage is particularly evident when handling significant boundary changes and occlusions, where our inpainting models realistically complete both texture and background.

\clearpage
\section{Pseudo Code for Inpaint4Drag}
To complement the detailed description of our method of the main paper, we provide a concise algorithmic representation of the data flow in \cref{algo:method}.

\begin{algorithm}[h]
\caption{Inpaint4Drag: Drag-based Image Editing via Bidirectional Warping and Inpainting}
\label{algo:method}
\SetAlgoLined
\KwIn{Image $I$, user-drawn mask $M$, handle points $\{h_i\}$ and target positions $\{t_i\}$}
\KwOut{Edited image $I_{\text{edit}}$}

\textbf{Region Specification and Boundary Refinement:}\;
$P_s \leftarrow$ SampleGridPoints($M$) \tcp*{Sample grid points from user mask}
$M_{pred} \leftarrow f_{SAM}(I, P_s)$ \tcp*{SAM prediction}
$M_{dilated} \leftarrow$ Dilate($M$, $K_1$); $M_{eroded} \leftarrow$ Erode($M$, $K_1$) \tcp*{Create boundary constraints}
$M \leftarrow (M_{pred} \cap M_{dilated}) \cup M_{eroded}$ \tcp*{Boundary-guided refinement}

\textbf{Bidirectional Warping:}\;
$\mathcal{C} \leftarrow$ ExtractContours($M$) \tcp*{Get deformable regions}
\ForEach{contour $C \in \mathcal{C}$}{
    Associate control points: $(h_i, t_i) \leftarrow \{(h_i, t_i) \mid h_i \text{ inside } C\}$\;
    
    \textbf{Forward Warping:} \tcp*{Define target region boundary}
    \ForEach{point $p$ in $C$}{
        $w_i \leftarrow \frac{1/(\|p - h_i\| + \epsilon)}{\sum_{j} 1/(\|p - h_j\| + \epsilon)}$\;
        $p_t \leftarrow p + \sum_{i} w_i(t_i - h_i)$ \tcp*{Weighted interpolation}
        Store mapping pair $(p, p_t)$\;
    }
    $C' \leftarrow$ transformed contour from forward warping\;
    
    \textbf{Backward Mapping:} \tcp*{Ensure complete pixel coverage}
    \ForEach{pixel $p_t$ within boundary of $C'$}{
        Find $N_n$ nearest matched pixels $\{p_i^{tgt}\}$ with source positions $\{p_i^{src}\}$\;
        $p_s \leftarrow p_t + \sum_{i=1}^{N_n} w_i(p_i^{src} - p_i^{tgt})$ \tcp*{Local neighborhood interpolation}
        \If{$p_s \in [0,W) \times [0,H)$ and $p_t \in [0,W) \times [0,H)$}{
            Store valid mapping $(p_s, p_t)$\;
        }
    }
}

\textbf{Compute Warped Image and Inpainting Mask:}\;
\ForEach{valid mapping pair $(p_s, p_t)$}{
    $I_{\text{warped}}(p_t) \leftarrow I(p_s)$ \tcp*{Transfer pixel values}
}
$M_{\text{warped}} \leftarrow$ mask of pixels filled in warped image\;
$M_{\text{temp}} \leftarrow M \setminus M_{\text{warped}}$\;
$\partial M_{\text{warped}} \leftarrow$ boundary of $M_{\text{warped}}$ \tcp*{Identify unmapped regions}
$M_{\text{inpaint}} \leftarrow$ Dilate$(M_{\text{temp}} \cup \partial M_{\text{warped}}, K_2)$ \tcp*{Create buffer zone}

\textbf{Image Inpainting:}\;
$I_{\text{edit}} \leftarrow$ Inpaint$(I_{\text{warped}}, M_{\text{inpaint}})$ \tcp*{Apply inpainting model}

\Return{$I_{\text{edit}}$}
\end{algorithm}

\clearpage

\section{Limitations}
While our method achieves significant improvements in efficiency and precision, it relies on accurate user-specified masks and control points for optimal performance. Imprecise user inputs, such as masks that inadvertently include background elements or poorly positioned control points, can lead to undesired deformation artifacts. Future work could explore understanding-enabled models that automatically filter irrelevant background elements or provide intelligent suggestions for mask and control point placement, reducing the burden on users to provide perfectly accurate inputs while maintaining the intuitive nature of drag-based interaction.

\end{document}

%% file: preamble.tex
\usepackage[ruled,vlined]{algorithm2e}
\usepackage{amsmath}
\usepackage{amssymb}

%% file: sec/0_abstract.tex
\begin{abstract}
Drag-based image editing has emerged as a powerful paradigm for intuitive image manipulation. However, existing approaches predominantly rely on manipulating the latent space of generative models, leading to limited precision, delayed feedback, and model-specific constraints. Accordingly, we present Inpaint4Drag, a novel framework that decomposes drag-based editing into pixel-space bidirectional warping and image inpainting. Inspired by elastic object deformation in the physical world, we treat image regions as deformable materials that maintain natural shape under user manipulation. Our method achieves real-time warping previews (0.01s) and efficient inpainting (0.3s) at 512×512 resolution, significantly improving the interaction experience compared to existing methods that require minutes per edit. By transforming drag inputs directly into standard inpainting formats, our approach serves as a universal adapter for any inpainting model without architecture modification, automatically inheriting all future improvements in inpainting technology. Extensive experiments demonstrate that our method achieves superior visual quality and precise control while maintaining real-time performance. \textcolor{black}{Project page: \url{https://visual-ai.github.io/inpaint4drag}}
\end{abstract}

%% file: sec/1_intro.tex
\section{Introduction}
\label{sec:intro}
\input{figs/teaser}
Image manipulation remains a fundamental challenge in computer vision, with increasing demand for intuitive tools that enable users to naturally modify image content. Among various interaction paradigms, drag-based editing has emerged as a promising direction \cite{pan2023drag, shi2023dragdiffusion}, offering an intuitive way to directly manipulate image elements through simple mouse operations. Recent advances in this field have demonstrated impressive results \cite{lu2024regiondrag, zhao2024fastdrag}, allowing users to move, resize, or deform objects within images through simple drag operations.

However, existing drag-based editing methods face several fundamental limitations. These approaches \cite{hou2024easydrag, mou2023dragondiffusion, ling2023freedrag, pan2023drag} rely on manipulating the latent space of generative models like Stable Diffusion \cite{nie2023blessing, luo2024readout, rombach2022high}, leading to three key challenges: (1) imprecise control -- latent space manipulations obscure the relationship between user inputs and resulting changes when control points are downscaled to match the lower latent resolution (\eg, from 512×512 to 32×32 in SD-UNet), (2) poor interactivity -- users are forced into time-consuming trial-and-error cycles without immediate visual feedback during the generative process, and (3) limited capability (shown in \cref{fig:teaser}) -- these approaches often produce unrealistic results when handling large occlusions (\eg, rotating heads or opening mouths) as they rely on general-purpose text-to-image models that are not specifically trained to handle missing or occluded regions.

In this paper, we present Inpaint4Drag, a novel interactive framework that decomposes drag-based image editing into bidirectional warping and image inpainting. Inspired by elastic object deformation in the physical world \cite{sederberg1986free, schaefer2006image}, we treat image regions as deformable materials that maintain natural shape under user manipulation. Given an input image, users specify a region mask to define the deformable area and place handle and target control points to guide the transformation. To assist with complex object manipulation, we provide an \textit{optional} mask refinement module that automatically captures precise object boundaries, ensuring consistent deformation of both edges and interior regions. The core of our approach lies in the bidirectional warping algorithm that transforms the input region based on user control inputs. While forward warping alone moves source pixels to their target positions, it inevitably creates holes and gaps that lead to artifacts. We address this by combining forward warping to define initial contours and rough deformation, with backward warping to fill gaps and establish dense correspondence for the geometric transformation. The warped result is complemented by computed inpainting masks that include dilated revealed regions (areas that emerge after image deformation) and a narrow band around warped contours for boundary smoothing. This complete input is then passed to modern inpainting models (\eg, LaMa \cite{suvorov2022resolution}, SD-Inpaint \cite{rombach2022high}) for final completion. 
Our decomposition enables a clear separation between geometric transformation and content generation while maintaining the familiar drag interaction paradigm.

Our physics-inspired formulation and bidirectional warping algorithm enable effective drag-based manipulation for inpainting models. It possesses three key advantages: First, our pixel-space deformation estimation enables precise geometric control while preserving colors of dragged content and maintaining image quality in unedited regions. Second, our method effectively handles large occlusions by leveraging specialized inpainting models, enabling challenging edits like opening a lion's mouth or rotating a person's head. Third, our approach serves as a universal adapter for the inpainting field - by transforming drag inputs directly into standard inpainting formats, we enable any inpainting model to function as a drag editing method without architecture modification, automatically inheriting all future improvements in inpainting technology. Notably, our system achieves exceptional interaction fluidity with real-time warping previews (0.01s) and efficient inpainting operations (0.3s) (measured on 512×512 images). 
We summarize our technical contributions as follows:
\begin{itemize}
\item A physics-inspired deformation framework that treats image regions as elastic materials, enabling natural transformations through user-specified control points and region masks, with \textit{optional} mask refinement for precise object boundary handling.

\item An efficient bidirectional warping algorithm that establishes initial shape through forward warping and fills gaps via backward mapping, creating dense pixel correspondences while maintaining real-time performance.

\item A modular pipeline that clearly separates transformation from generation, computing precise masks for revealed regions and boundary smoothing to enable seamless integration with existing inpainting models.
\end{itemize}

%% file: figs/teaser.tex
\begin{figure*}[t]
\centering
\includegraphics[width=\textwidth]{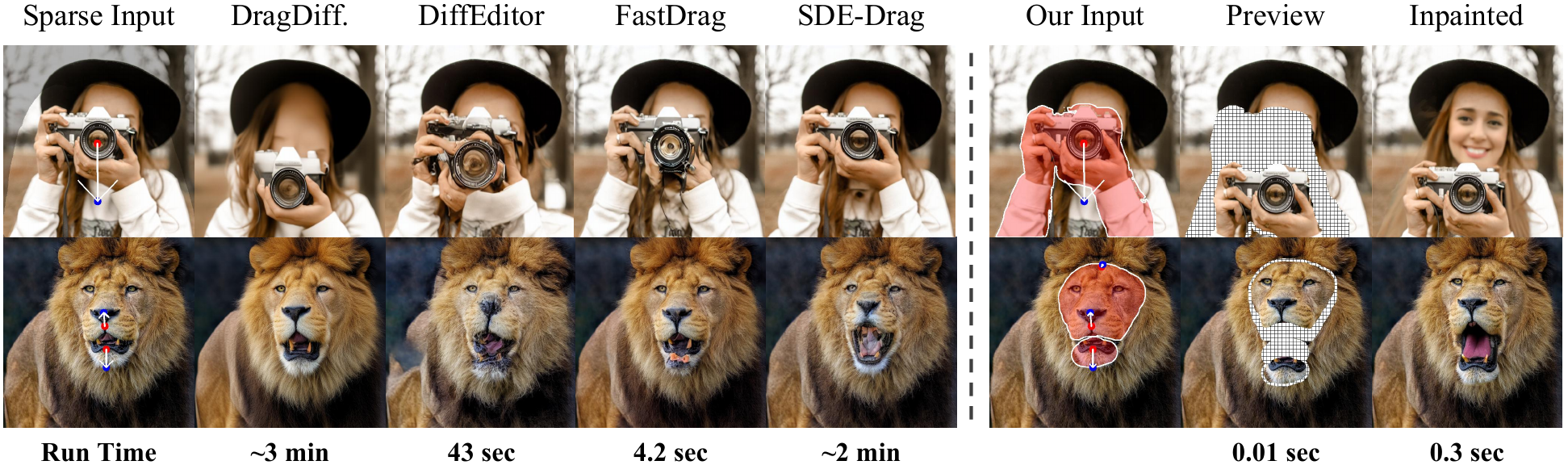}
\caption{Comparison of drag-based image editing methods. Our approach (rightmost columns) enables both precise edits and large-scale occlusion handling with real-time preview (0.01 sec) followed by rapid inpainting (0.3 sec). Users select \textcolor{red}{deformable regions} and drag from \textcolor{red}{handle points} to \textcolor{blue}{target positions}, with the preview column showing grid overlays in areas requiring inpainting. In contrast, existing methods (leftmost columns) require substantially longer processing times without providing interactive feedback during editing, and often struggle with precise manipulations due to latent-space operations.}
\label{fig:teaser}
\vspace{-4mm}
\end{figure*}

%% file: sec/2_related_work.tex
\section{Related Work}
\label{sec:related}
\noindent \textbf{Generative Models for Image Editing.}
Generative models have revolutionized image editing, with generative adversarial networks (GANs) marking a pivotal breakthrough \cite{goodfellow2014generative, karras2020analyzing, karras2021alias, kang2023scaling}. While numerous editing techniques have emerged from the GAN framework \cite{endo2022user, patashnik2021styleclip, xia2022gan, wang2022high, creswell2018inverting, yang2021gan, pan2023drag}, their practical applications remain constrained by training data diversity, model capacity, and challenges in inverting real images into GAN latent spaces \cite{creswell2018inverting, zhu2016generative, li2020gan}.
The emergence of large-scale text-to-image diffusion models \cite{rombach2022high, saharia2022photorealistic} has expanded image manipulation capabilities through text prompts \cite{cao2023masactrl, brooks2023instructpix2pix, epstein2024diffusion, kawar2023imagic, avrahami2023blended}, enabling control over style, motion, and object categories. However, these text-based approaches lack pixel-level precision, focusing primarily on semantic alterations.
To address this limitation, drag-based techniques \cite{endo2022user, pan2023drag, ling2023freedrag, shi2023dragdiffusion, luo2024rotationdrag, mou2023dragondiffusion, mou2024diffeditor} have emerged, allowing fine-grained manipulation of object posture and shape through interactive keypoint control, effectively combining generative modeling with the precision needed for detailed image editing.

\noindent \textbf{Drag-based Image Editing.}
Drag-based methods have transformed image editing by enabling feature manipulation through paired handle and target points. DragGAN \cite{pan2023drag} pioneered multi-point editing in GAN-generated images using iterative motion supervision and point tracking, while FreeDrag \cite{ling2023freedrag} enhanced precision by operating in feature space.
Building on these foundations, several diffusion-based approaches have emerged. DragDiffusion \cite{shi2023dragdiffusion} adapted the supervise-and-track framework to diffusion models, while DiffEditor \cite{mou2024diffeditor} and EasyDrag \cite{hou2024easydrag} extended feature dragging across the denoising process. SDE-Drag \cite{nie2023blessing} introduced sparse latent copy-paste techniques, and RegionDrag \cite{lu2024regiondrag} improved efficiency through dense region mapping. InstantDrag \cite{shin2024instantdrag}, a motion prediction approach, combines motion guidance with generative models but requires test-time optimization on video clips for each new edit. FastDrag \cite{zhao2024fastdrag}, while sharing our intuition of object stretching, operates solely in latent space with sequential processing, lacking our method's vectorized pixel-space manipulation and specialized inpainting capabilities.

\noindent \textbf{Image Inpainting.}
Image inpainting has evolved from classical PDE-based~\cite{bertalmio2000image} and patch-based~\cite{barnes2009patchmatch,criminisi2004region} methods to modern deep learning approaches. CNN-based methods~\cite{pathak2016context} first learned semantic priors, followed by GANs~\cite{yu2018generative,liu2018image} and transformers~\cite{li2022mat} that improve the inpainting performance through adversarial training or enhancing long-range consistency. Recent diffusion models like Stable Diffusion~\cite{rombach2022high} and LaMa~\cite{suvorov2022resolution} achieve superior quality in structural coherence and detail synthesis, widely adopted by commercial tools like Adobe Photoshop~\cite{adobe2024photoshop}, DALL-E3~\cite{openai2024dalle}, and RunwayML~\cite{runway2024ml}, which inspires us to consider the emerging drag-based image editing problem from the inpainting perspective.

%% file: sec/3_method.tex
\input{figs/method_1}

\section{Method}
\label{sec:method}
We present Inpaint4Drag, an interactive approach for drag-based image editing that decomposes the task into bidirectional warping and standard image inpainting. Given an input image, users can draw masks to select objects for deformation, with our mask refinement module to improve boundary precision (\cref{sec:user_input}). After specifying drag point pairs, our bidirectional warping algorithm immediately deforms the selected region (\cref{sec:bidirectional_warping}), providing real-time preview of the editing result that also serves as input for the inpainting model. Users iteratively refine masks and control points to achieve desired dragging effects before executing the relatively expensive inpainting operation, which fills regions revealed by deformation (\cref{sec:inpainting}). Our approach enables real-time interaction while delivering high-quality results through specialized deformation control and integrated inpainting models for seamless completion.

\subsection{Region Specification and Boundary Refinement}
\label{sec:user_input}

Previous drag editing methods \cite{pan2023drag, shi2024instadrag, zhao2024fastdrag} typically use sparse control points to guide deformation, with optional masking to restrict editable regions. However, this sparse input format introduces fundamental ambiguity in deformation interpretation -- the movement of numerous pixels relies on the guidance from only a few control points. A single drag operation on a character's arm could produce vastly different outcomes: full-body rotation, localized arm movement, or isolated point translation (see S4 in supplementary material for examples). Without clear deformation specifications, existing methods often produce unpredictable results that deviate from user intentions.

We address this ambiguity by requiring users to explicitly mask regions $M$ intended for deformation, similar to how we naturally identify movable parts of objects. Given an input image $I \in \mathbb{R}^{H \times W \times 3}$, a set of handle points $\{h_i\}$ and their target positions $\{t_i\}$, our key insight is to treat each masked region as an elastic material, where moving a handle point creates a ripple effect throughout the connected area -- much like stretching a rubber sheet. Our bidirectional warping approach (\cref{sec:bidirectional_warping}) computes dense deformation fields that ensure smooth influence propagation from control points to every pixel $p \in M$ while maintaining complete pixel coverage throughout the deformed region.
\input{figs/method_2}
For effective deformation propagation, mask boundaries should align with object boundaries -- pixels belonging to the same object should move coherently. To maintain deformation coherence across object boundaries while simplifying user interactions, we propose an \textit{optional} mask refinement module based on the Segment Anything Model (SAM) \cite{kirillov2023segment}. Given a user input mask $M$ containing points $P = \{p_i\}_{i=1}^N$, directly using all mask points as SAM input would introduce computational bottlenecks. To address this, we sample grid points $P_s \subset P$ from the user-drawn mask as SAM input, which maintains interactive performance while preserving critical boundary information.

We process these sampled points through SAM's prompt encoder $f_\text{SAM}$ to obtain point embeddings, which are combined with the image embedding to generate a prediction mask $M_\text{pred} = f_\text{SAM}(I, P_s)$. Despite SAM's powerful segmentation capabilities, we found that direct SAM predictions can generate disconnected regions or capture unintended objects far from the user's specified area due to SAM's tendency to segment complete objects and semantically similar instances rather than user-intended regions (see \cref{fig:method_1} or \cref{fig:mask_refinement_series} for examples). We address this limitation with a two-step refinement approach that balances automatic boundary detection while preserving user intent. First, we generate dilated and eroded versions of the input mask:
\begin{equation}
    M_\text{dilated}(x,y) = \max_{(i,j) \in K_1} M(x-i, y-j),
\end{equation}
\begin{equation}
M_\text{eroded}(x,y) = \min_{(i,j) \in K_1} M(x-i, y-j),
\end{equation}
where $K_1$ represents a dilation / erosion kernel with radius $r_1$. The refined mask $M$ is obtained through our boundary-guided formulation:
\begin{equation}
M = (M_\text{pred} \cap M_\text{dilated}) \cup M_\text{eroded},
\end{equation}
where $\cup$ and $\cap$ represent the logical OR and AND operations respectively. This approach uses the dilated and eroded masks to create natural boundary constraints, ensuring the refined result respects the user's original intent while improving boundary precision. By exposing the kernel radius $r_1$ as a control parameter, users can tailor the refinement strength, achieving their desired balance between automatic boundary detection and manual control.

\subsection{Bidirectional Warping for Region Deformation}
\label{sec:bidirectional_warping}

Given a user mask $M$ (either original or refined) and control point pairs $\{(h_i, t_i)\}_{i=1}^K$, our bidirectional warping algorithm aims to translate these inputs into a coherent deformation of the masked region and prepares standard inputs for inpainting models: a warped image $I_{\text{warped}}$ containing deformed content alongside an inpainting mask $M_{\text{inpaint}}$ identifying areas left vacant by relocated pixels.

As shown in \cref{fig:method_2} and \cref{fig:method_3}, our approach consists of four steps: (1) contour extraction to identify independent deformable regions; (2) forward warping to define target region boundaries and establish initial mapping; (3) backward mapping to ensure complete pixel coverage in target regions; and (4) leveraging the established mapping to generate warped content and identify areas requiring inpainting. 

\subsubsection{Contour Extraction and Control Point Association}
We first decompose the binary mask $M$ into distinct contours representing separate deformable regions (left of \textcolor{myPink}{pink block}, \cref{fig:method_2}). A contour here refers to a closed curve that traces the boundary of a connected area in the mask:
\begin{equation}
\mathcal{C} = \text{findContours}(M).
\end{equation}
To maintain local deformation coherence, we associate each contour point $c \in \mathcal{C}$ with control points located inside its region:
\begin{equation}
(H_c, T_c) = \{(h_i, t_i) \, | \, h_i \text{ inside } \mathcal{C}\}.
\end{equation}
This ensures that each region responds only to control points within its boundaries, preserving the intuitive behavior where deformation occurs only where user interaction is directly applied. For brevity, we simply use $(h_i, t_i)$ in subsequent equations to refer to a control points associated with a contour point $c$.
\input{figs/method_3}
\subsubsection{Forward Warping}
For regions with a single control point pair, deformation simplifies to translation, where region pixels directly shift from handle to target position. When multiple control points are present, the deformation becomes non-rigid, requiring a interpolation-based transformation. We first perform forward warping (\textcolor{myGreen}{green block}, \cref{fig:method_2})  which serves two critical purposes: 
i) it transforms the contour itself to define the target region boundary $\mathcal{C}'$, and 
ii) it establishes initial pixel-level mapping to guide subsequent processing. 
For each point $p$ in the source region (including boundary points of contour $\mathcal{C}$), we compute its target position $p_t$ through weighted interpolation:
\begin{equation}
p_t = p + \sum_{i=1}^{N_\mathcal{C}} w_i(t_i - h_i),
\end{equation}
where $N_\mathcal{C}$ denotes the number of control points associated with contour $\mathcal{C}$, and weights $w_i$ are computed through inverse distance weighting:
\begin{equation}
w_i = \frac{1/(\|p - h_i\| + \epsilon)}{\sum_{j=1}^{N_C} 1/(\|p - h_j\| + \epsilon)}.
\end{equation}

While this forward approach establishes initial mapping, using it alone for warping creates sampling artifacts (see \cref{fig:method_2} or \cref{fig:bi} for examples). Specifically, when non-rigid transformations stretch image regions, the discrete nature of pixels creates gaps where target locations receive no mapped value. This occurs when the transformed source pixel grid becomes stretched and discontinuous in the target space, leaving unmapped gaps.

\subsubsection{Backward Mapping}
To ensure complete pixel coverage within the deformed region (\textcolor{blue}{blue block}, \cref{fig:method_2}),  we compute source positions for all pixels in the transformed target contour $C'$. For each target pixel $p_t \in C'$, we determine its corresponding source position $p_s$ using:
\begin{equation}
p_s = p_t + \sum_{i=1}^{N_n} w_i(p_i^\text{src} - p_i^\text{tgt}),
\end{equation}
where $N_n$ represents the $n$ nearest target pixels that were successfully matched during the forward process, $p_i^\text{tgt}$ represents one of these matched nearest target positions, $p_i^\text{src}$ is its corresponding source position, and $w_i$ are the inverse distance weights as previously defined.

This local neighborhood approach provides computational efficiency compared to using global pixel references and helps preserve structural coherence by limiting the influence of distant forward warping results on the final mapping. We validate each computed mapping pair by ensuring all coordinates remain within image boundaries:
\begin{equation}
\text{Valid}(p_s, p_t) = p_s, p_t \in [0,W) \times [0,H),
\end{equation}
where $W$ and $H$ denote the image width and height, respectively. This prevents sampling from undefined regions outside the image domain. 

Overall, our backward mapping strategy completes the bidirectional framework by establishing reliable pixel-level mappings that maintain local structural relationships while addressing the limitations of forward transformation.

\subsubsection{Computing Warped Result and Inpainting Mask}

Using the established pixel mappings, we generate the warped image by transferring pixel values:
\begin{equation}
I_{\text{warped}}(p_t) = I(p_s) \text{ for all valid } (p_s, p_t)~\text{pairs}.
\end{equation}
As shown in \cref{fig:method_3}, we then identify regions requiring inpainting by determining pixels present in the original mask $M$ but unmapped in the deformed result:
\begin{equation}
M_{\text{temp}} = M \setminus M_{\text{warped}}.
\end{equation}
To ensure smooth transitions at both warped content boundaries and vacated regions, we apply dilation to both the unmapped areas $M_{\text{temp}}$ and the boundaries of the warped mask $\partial M_{\text{warped}}$. This creates a buffer zone around the boundaries by expanding the inpainting mask, allowing the inpainting model to handle the transition areas and avoid abrupt edges between warped and newly generated content:
\begin{equation}
M_{\text{inpaint}} = \text{dilate}(M_{\text{temp}} \cup \partial M_{\text{warped}}, K_2),
\end{equation}
where $K_2$ represents the dilation kernel with radius $r_2$. To this end, our bidirectional warping algorithm has processed the user mask $M$ and control point pairs $\{(h_i, t_i)\}_{i=1}^K$ to produce the warped image $I_{\text{warped}}$ and an inpainting mask $M_{\text{inpaint}}$, which together form the standard input required by image inpainting models.

\input{figs/qualitative}

\subsection{Integration with Image Inpainting}
\label{sec:inpainting}

The final step in our pipeline is to apply an inpainting model to generate content for areas revealed during deformation:
\begin{equation}
I_{\text{edit}} = \text{Inpaint}(I_{\text{warped}}, M_{\text{inpaint}}).
\end{equation}

Before executing inpainting operations, our method provides $I_{\text{warped}}$ as a real-time preview of the final result $I_{\text{edit}}$—a capability absent in existing approaches. This preview enables users to achieve desired dragging effects by adjusting mask and control points, resulting in a more interactive editing experience that avoids unnecessary expense from repeated inpainting attempts.

For our implementation, we selected the Stable Diffusion 1.5 Inpainting Checkpoint \cite{rombach2022high}, which was fine-tuned from the regular Stable Diffusion v1.2 model with additional training for inpainting tasks. The inpainting process follows a straightforward pipeline: the mask is resized and concatenated with the image VAE latent representation. During conditional diffusion denoising, we initialize masked regions with pure noise to generate entirely new content in areas revealed by deformation. Finally, the result is transformed back to pixel space through VAE decoding.

To optimize performance, we incorporated several efficiency enhancements: TinyAutoencoder SD (TAESD) \cite{ollin2023taesd}, a distilled VAE that reduces memory requirements; LCM (Latent Consistency Model) LoRA \cite{luo2023lcm} to reduce sampling steps; an empty text prompt to eliminate classifier-free guidance computation; and caching of the empty prompt embeddings to avoid repetitive calculations during editing sessions. It's worth noting that while we report experimental results using this representative inpainting model, our framework can accommodate any inpainting model as a drop-in replacement (see S2 in supplementary material for examples).


%% file: figs/method_1.tex
\begin{figure}[t]
    \centering
    \includegraphics[width=0.48\textwidth]{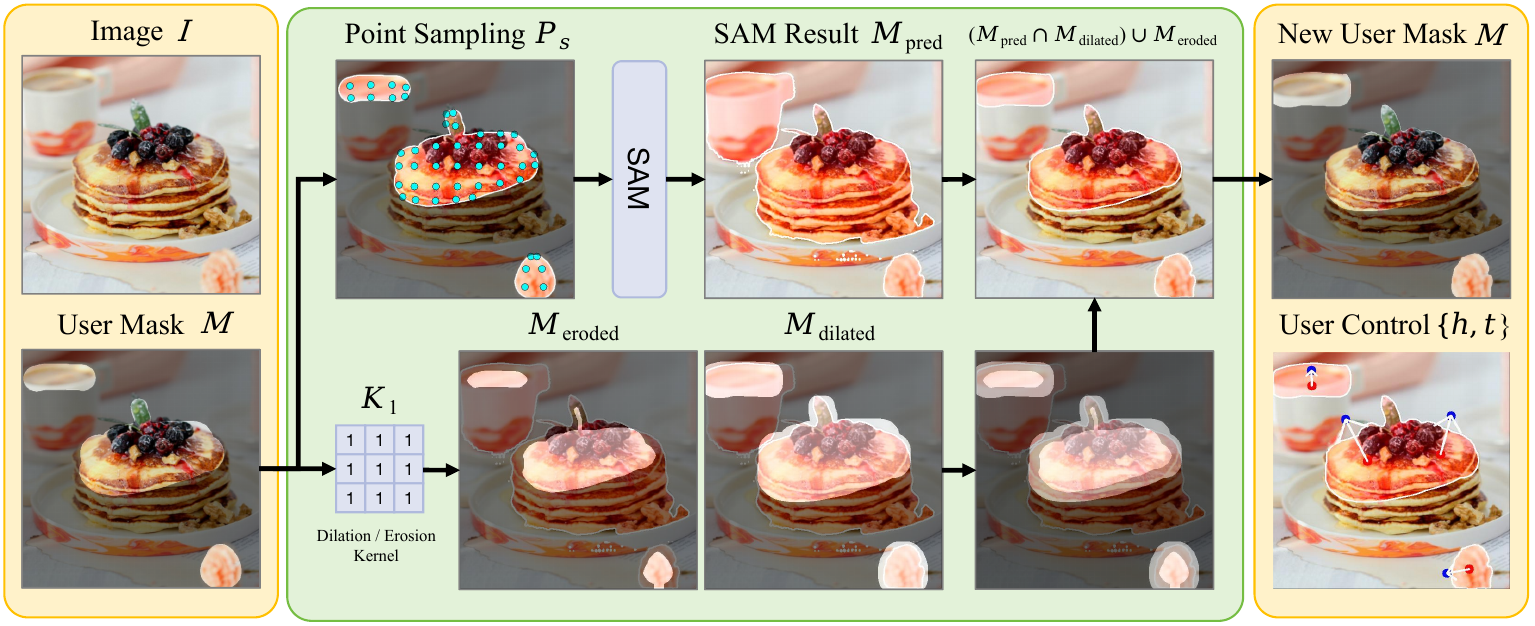}
    \caption{Overview of our \textit{optional} mask refinement module. Users can achieve precise object boundaries for coherent deformation through SAM \cite{kirillov2023segment}, constrained by eroded and dilated mask boundaries to preserve user intent.}
    \label{fig:method_1}
    \vspace{-4mm}
\end{figure}

%% file: figs/method_2.tex
\begin{figure*}[t]
    \centering
    \includegraphics[width=\textwidth]{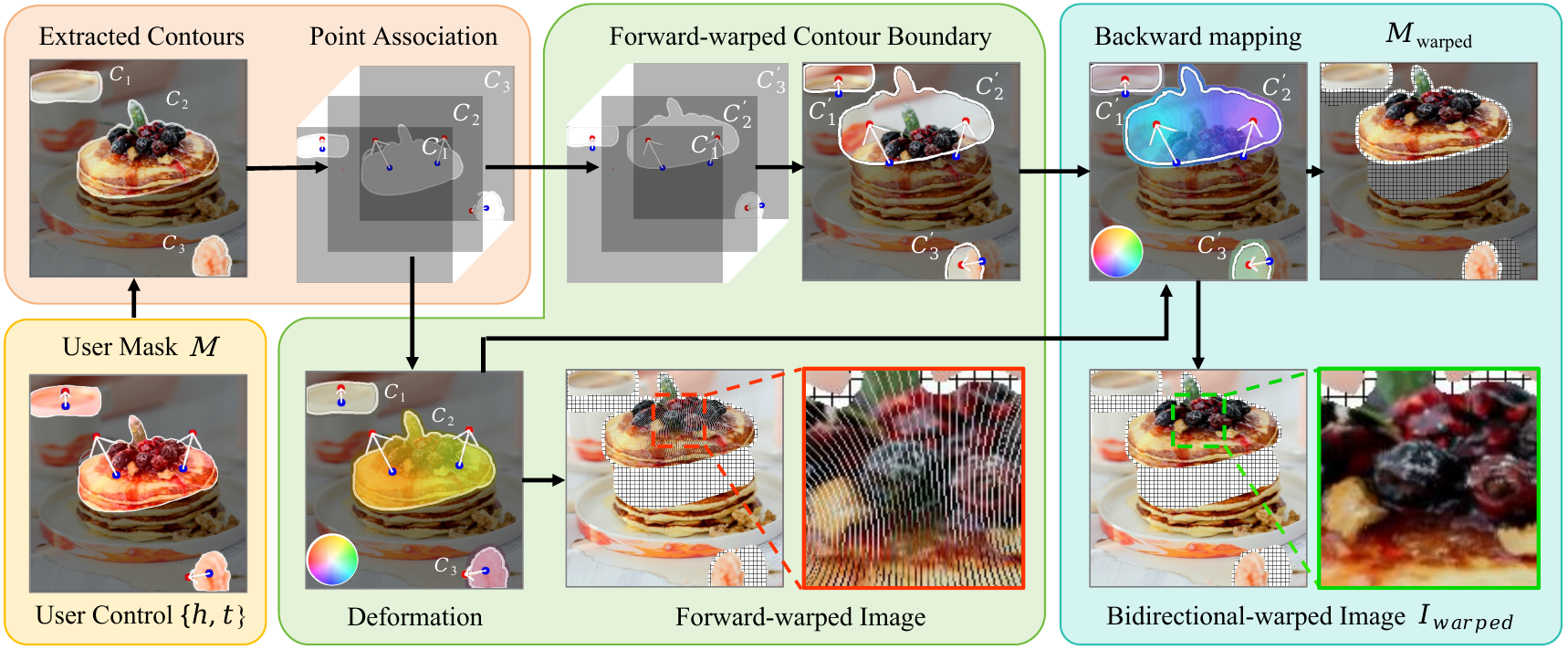}
    \caption{Overview of our bidirectional warping pipeline. Given a user mask and control points, we first extract region contours and establish point associations. The forward warping step then maps these contours to their target locations and builds initial correspondences. Finally, through backward mapping, we generate the warped image with complete pixel coverage and its corresponding warped mask, which will be used for subsequent inpainting.}
    \label{fig:method_2}
    \vspace{-4mm}
\end{figure*}

%% file: figs/method_3.tex
\begin{figure*}[t]
    \centering
    \includegraphics[width=\textwidth]{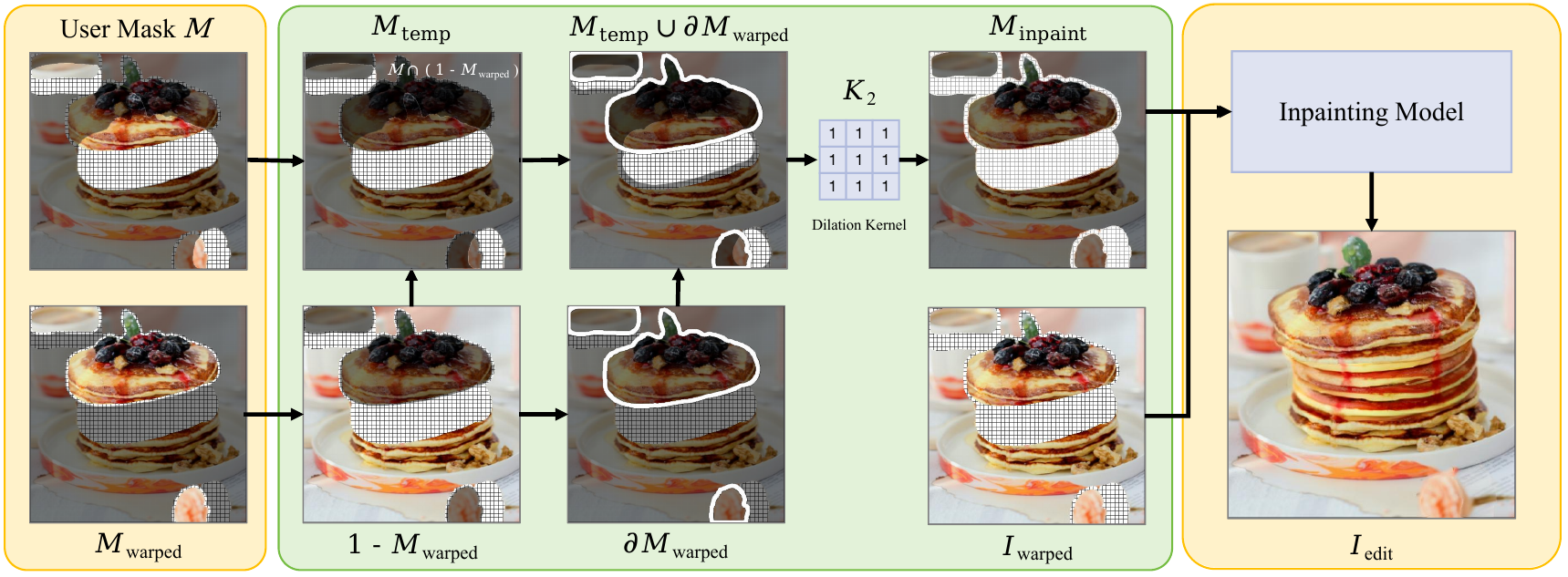}
    \caption{Overview of inpainting mask computation. The final mask ($M_{\text{inpaint}}$) combines dilated unmapped regions ($M_{\text{temp}}$) from areas absent after warping and dilated boundaries of the warped mask ($\partial M_{\text{warped}}$). This ensures smooth transitions between warped and generated regions in the final result.}
    \label{fig:method_3}
    \vspace{-4mm}
\end{figure*}

%% file: figs/qualitative.tex
\begin{figure*}[ht]
    \centering
    \includegraphics[width=\linewidth]{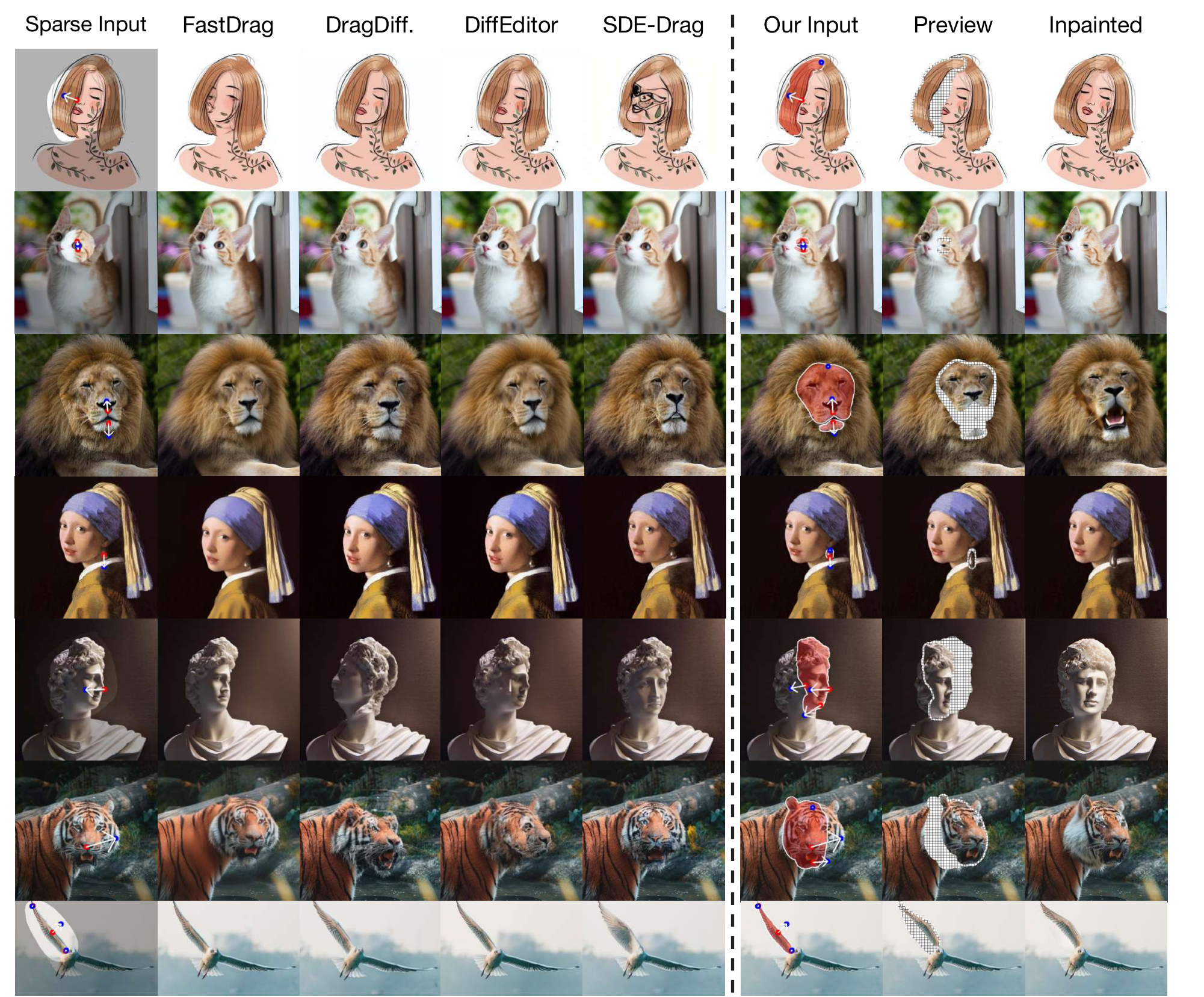}
    \caption{Qualitative comparison with state-of-the-art methods on challenging cases from DragBench-S \cite{nie2023blessing} and DragBench-D \cite{shi2023dragdiffusion}. Our method enables users to specify \textcolor{red}{deformable regions} and manipulate them by dragging \textcolor{red}{handle points} toward \textcolor{blue}{target destinations}. The grid overlay in the preview column indicates areas to be inpainted. Our approach shows advantages over existing methods by effectively preserving local details in dragged regions through our bidirectional warping algorithm while demonstrating strong capability in generating occluded regions. The real-time preview ($\sim$10ms) allows users to interactively adjust edits before executing the inpainting process ($\sim$0.3s).}
    \label{fig:qualitative}
\end{figure*}

%% file: sec/4_experiments.tex
\section{Experiments}
\subsection{Datasets}
We evaluate drag editing methods using two benchmarks: DragBench-S from SDE-Drag \cite{nie2023blessing} with 100 samples, and DragBench-D from DragDiffusion \cite{shi2023dragdiffusion} containing 205 samples. Each benchmark entry includes a source image, a text prompt describing the image, a binary mask for the editable region, and point pairs showing desired movements. In our framework, masks serve a different purpose as deformable regions. While editable regions indicate where editing is allowed, deformable regions define coherent parts that should move together during transformation. We therefore re-annotated the deformable region locations and dragging points while preserving the original user editing intentions and keeping the source images unchanged.

\subsection{Evaluation Metrics}
Following \cite{lu2024regiondrag}, we measure editing performance using LPIPS \cite{zhang2018perceptual} and Mean Distance (MD).

\noindent\textbf{LPIPS:} Learned Perceptual Image Patch Similarity (LPIPS) v0.1 \cite{zhang2018perceptual} measures identity preservation between original and edited images. Lower LPIPS indicates better identity preservation. However, a limitation of this metric is that it only measures low-level feature similarities between image patches. In drag editing, intentional and correct shape deformations often result in unavoidable LPIPS increases that should not be penalized.

\noindent\textbf{Mean Distance (MD):} This metric evaluates how accurately handle points are moved to target positions. DIFT \cite{tang2024emergent} is employed to find matched points for the user-specified handle points in the edited image, restricting the search area to regions around user-specified handle and target points to avoid false matches. MD is calculated as the average normalized Euclidean distance between target points and DIFT-matched points.

\subsection{Implementation Details}
Our framework is implemented in Python using PyTorch~\cite{paszke2019pytorch} on a single NVIDIA Tesla V100-SXM2 GPU. For mask refinement, we adopt EfficientVIT-SAM-L0~\cite{zhang2024efficientvit} and sample points from user masks, limiting the number of sampled points to $|P_s| \leq 128$ for computational efficiency. The dilation/erosion kernel radius $r_1$ is set to 10 pixels to provide sufficient boundary refinement while preserving user intent. In the bidirectional warping module, we set $\epsilon=10^{-6}$ for numerical stability in weight computation and use $N_n=4$ nearest neighbors for backward mapping. The final inpainting mask is dilated with kernel radius $r_2=5$ pixels to ensure smooth transitions. For inpainting, we employ Stable Diffusion 1.5 inpainting checkpoint~\cite{rombach2022high} with TAESD~\cite{ollin2023taesd} and LCM LoRA~\cite{luo2023lcm}, processing images resized to 512 pixels on the longer edge while maintaining aspect ratio. We use 8 sampling steps for diffusion without classifier-free guidance.

\input{figs/quantitative}
\subsection{Quantitative Evaluation}
As shown in \cref{tab: exp}, we compare Inpaint4Drag with state-of-the-art methods \cite{shi2023dragdiffusion,mou2024diffeditor,nie2023blessing,zhao2024fastdrag}. Our bidirectional warping algorithm enables significant improvements in dragging precision and image consistency on DragBench-S \cite{nie2023blessing} and DragBench-D \cite{shi2023dragdiffusion}, achieving the lowest MD scores (3.6/3.9) and competitive LPIPS values (11.4/9.1). This advantage stems from our dense pixel-wise deformation calculation that preserves color and geometric relationships during manipulation. Notably, Inpaint4Drag is 14× faster than FastDrag \cite{zhao2024fastdrag} and nearly 600× faster than DragDiffusion \cite{shi2023dragdiffusion}, with SAM-based refinement taking 0.02s and bidirectional warping only 0.01s. The computational peak occurs during SD inpainting for previews, requiring 0.29s while using the least memory (2.7GB) among all methods.

\subsection{Qualitative Results}
As shown in \cref{fig:qualitative}, Inpaint4Drag outperforms state-of-the-art methods on challenging cases from DragBench-S and DragBench-D (see S5 in supplementary materials for more results). By leveraging specialized inpainting models, Inpaint4Drag generates realistic content in previously occluded regions (\eg, newly exposed facial features in rows 1, and 4, and the lion's open mouth in row 3). Our \textit{optional} mask refinement module enables users to coherently deform object boundaries (rows 4, 5, and 7) or focus on precise local edits (remaining examples). Unlike latent-space methods that lose precision when downscaling control points to latent resolution, our pixel-space approach enables accurate manipulation of local details (row 2, and 5). Our bidirectional warping generates an informative preview of the dragged content, providing informative context for subsequent inpainting process.

\subsection{Method Analysis}
\input{figs/mask_fine}
\textbf{\noindent Mask Refinement.} As demonstrated in~\cref{fig:mask_refinement_series}, our approach transforms rough initial masks (left) into refined results (right) by constraining SAM's predictions (middle) within a dilated region of the user's input. While our refinement module incorporates both inner and outer boundary constraints, for visualization clarity, we only showcase the outer boundary usage in these examples. The resulting masks effectively capture object boundaries while preserving the user's intended editing scope.

\noindent\textbf{Unidirectional \textit{vs.} Bidirectional Warping.} 
We present qualitative comparisons between unidirectional (forward-warping-only) and bidirectional warping (\cref{sec:bidirectional_warping}) in \cref{fig:bi}. The unidirectional approach struggles with sampling artifacts, creating noticeable gaps in stretched regions during deformation. These artifacts emerge when non-rigid transformations stretch image regions, as the discrete nature of pixels results in unmapped gaps where target locations receive no source values due to discontinuities in the transformed pixel grid. Our bidirectional method effectively addresses these challenges through a two-step process: first identifying target contours via forward warping, then employing pixel-level backward mapping to fill the gaps. This approach yields smooth transformations without discontinuities, providing reliable visual context for both user preview and image inpainting.

\input{figs/uni_bi}

%% file: figs/quantitative.tex
\begin{table}[t]
\centering
\scriptsize
\begin{tabular}{@{\hspace{1pt}}l@{\hspace{4pt}}c@{\hspace{4pt}}c@{\hspace{4pt}}c@{\hspace{2pt}}c@{\hspace{4pt}}c@{\hspace{2pt}}c@{\hspace{1pt}}}
\toprule
Method & & & \multicolumn{2}{c}{DragBench-S} & \multicolumn{2}{c}{DragBench-D} \\
& Mem(GB)$\downarrow$ & Time(s)$\downarrow$ & MD$\downarrow$ & LPIPS$\downarrow$ & MD$\downarrow$ & LPIPS$\downarrow$ \\
\midrule
DragDiffusion \cite{shi2023dragdiffusion} & 11.6 & 177.7 & 7.0 & 18.0 & 6.7 & 10.2 \\
DiffEditor \cite{mou2024diffeditor} & 6.6 & 43.1 & 23.6 & 17.6 & 22.1 & 10.9 \\
SDE-Drag \cite{nie2023blessing} & 6.9 & 126.1 & 7.5 & \textbf{11.4} & 8.1 & 14.9 \\
FastDrag \cite{zhao2024fastdrag} & 5.0 & 4.2 & 4.1 & 24.1 & 5.1 & 13.5 \\
Ours & \textbf{2.7} & \textbf{0.3} & \textbf{3.6} & \textbf{11.4} & \textbf{3.9} & \textbf{9.1} \\
\bottomrule
\end{tabular}
\vspace{-2mm}
\caption{Comparison of different drag-based image editing methods. MD and LPIPS values are scaled by 100. Time and GPU memory are measured at 512$\times$512 resolution.}
\vspace{-6mm}
\label{tab: exp}
\end{table}

%% file: figs/mask_fine.tex
\begin{figure}[t]
    \centering
    \begin{minipage}{0.33\columnwidth}\centering Initial Mask\end{minipage}%
    \begin{minipage}{0.33\columnwidth}\centering SAM Prediction\end{minipage}%
    \begin{minipage}{0.33\columnwidth}\centering Refined Mask\end{minipage}
    \vspace{-4mm}
    \includegraphics[width=\columnwidth]{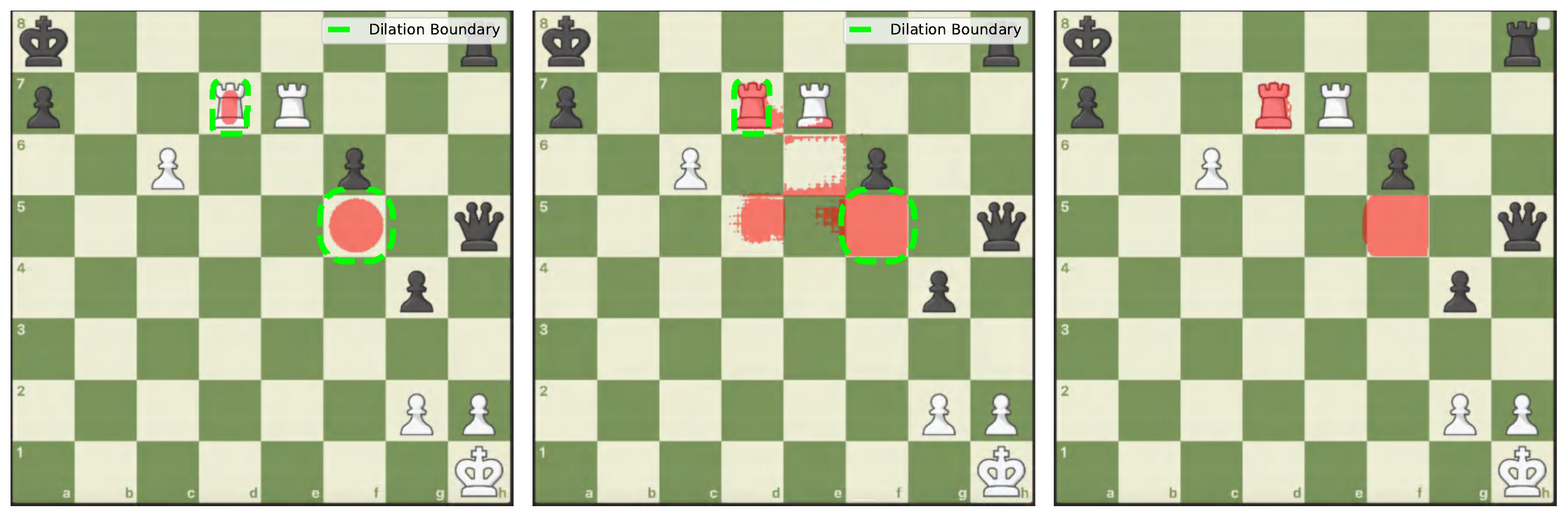}
    \includegraphics[width=\columnwidth]{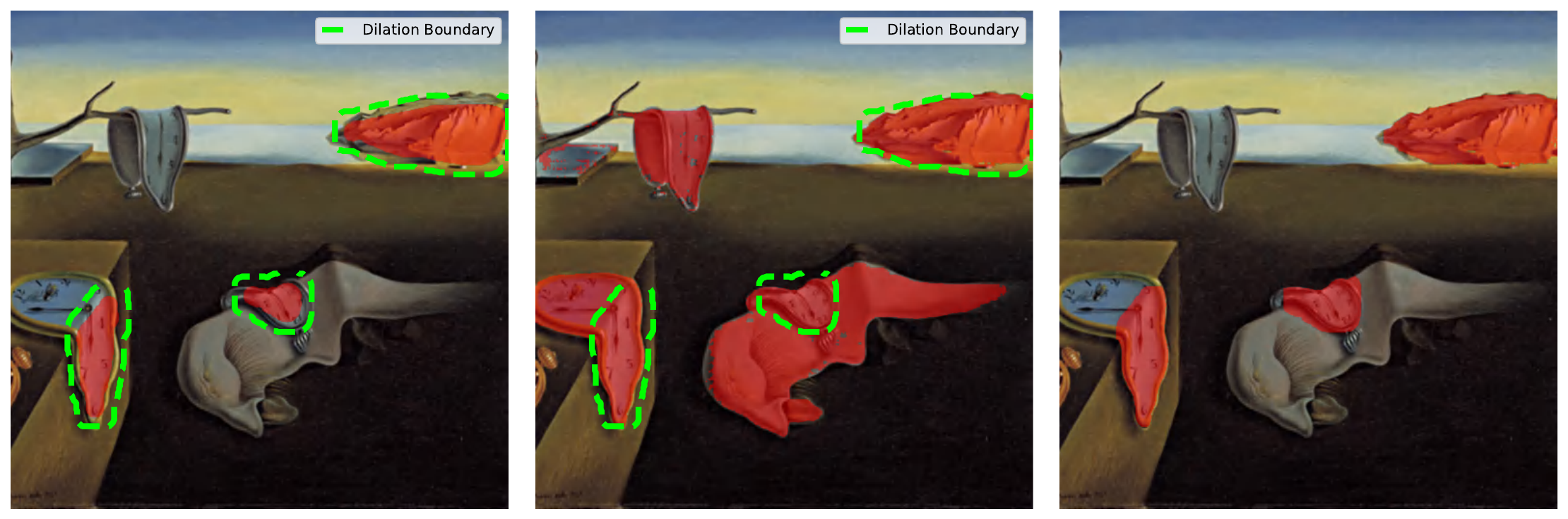}
    \caption{Qualitative results of our mask refinement module \cref{sec:user_input}. 
    \textbf{Left:} User-provided initial input mask. 
    \textbf{Middle:} Raw segmentation predictions from Segment Anything Model (SAM). 
    \textbf{Right:} Final refined results where the green dashed boundary, derived from dilating the initial input mask, effectively filters out undesired SAM predictions beyond the user's intended scope.}
    \label{fig:mask_refinement_series}
    \vspace{-2mm}
\end{figure}

%% file: figs/uni_bi.tex
\begin{figure}[t]
    \centering
    \begin{minipage}{0.32\columnwidth}\centering Input\end{minipage}%
    \hspace{1mm}%
    \begin{minipage}{0.32\columnwidth}\centering Unidirectional\end{minipage}%
    \hspace{1mm}%
    \begin{minipage}{0.32\columnwidth}\centering Bidirectional\end{minipage}

    \begin{tabular}{c@{\hspace{1mm}}c@{\hspace{1mm}}c}
        \includegraphics[width=0.32\columnwidth]{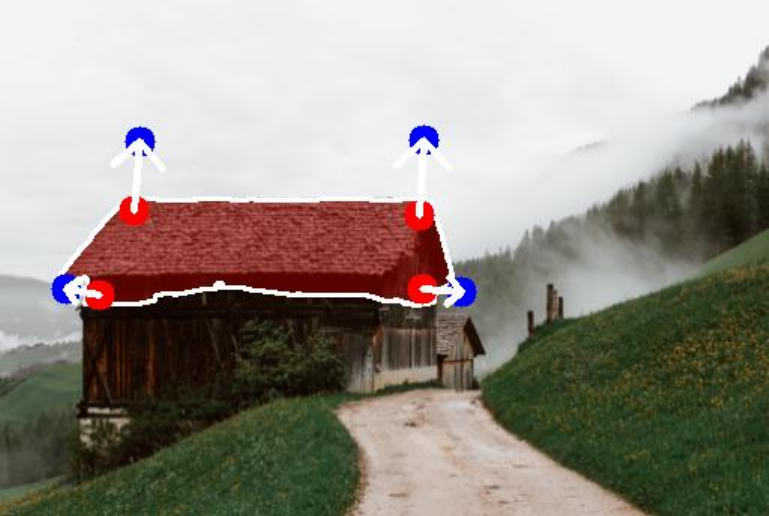} &
        \includegraphics[width=0.32\columnwidth]{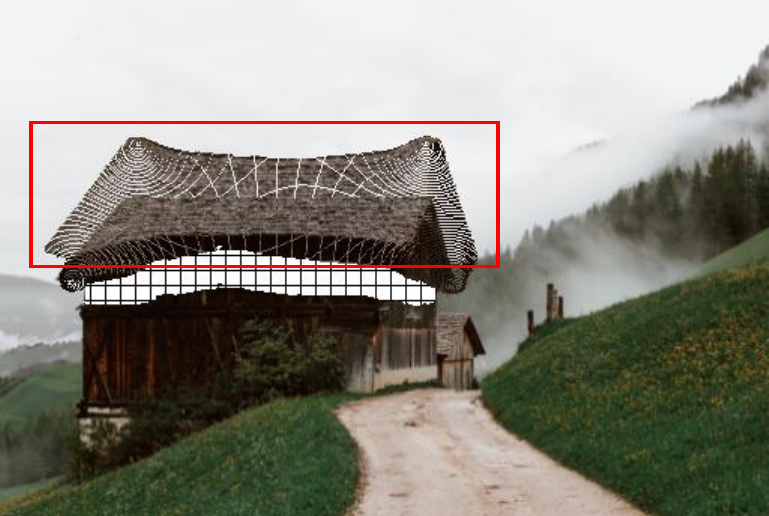} &
        \includegraphics[width=0.32\columnwidth]{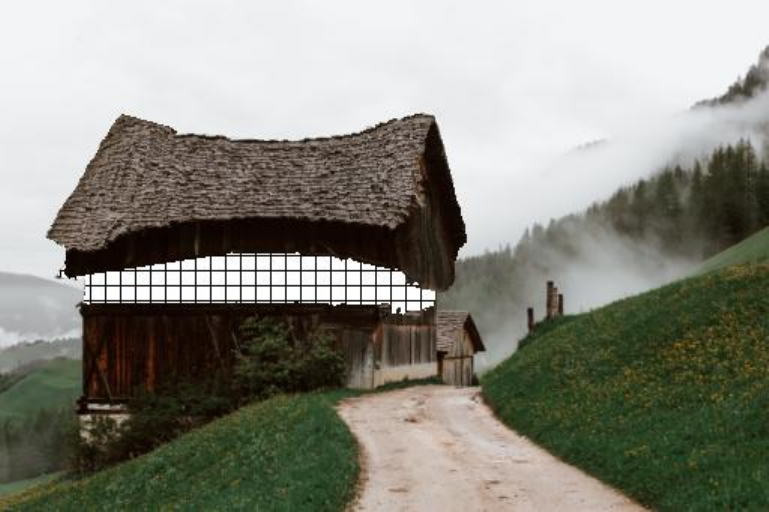} \\[-1ex]
        \includegraphics[width=0.32\columnwidth]{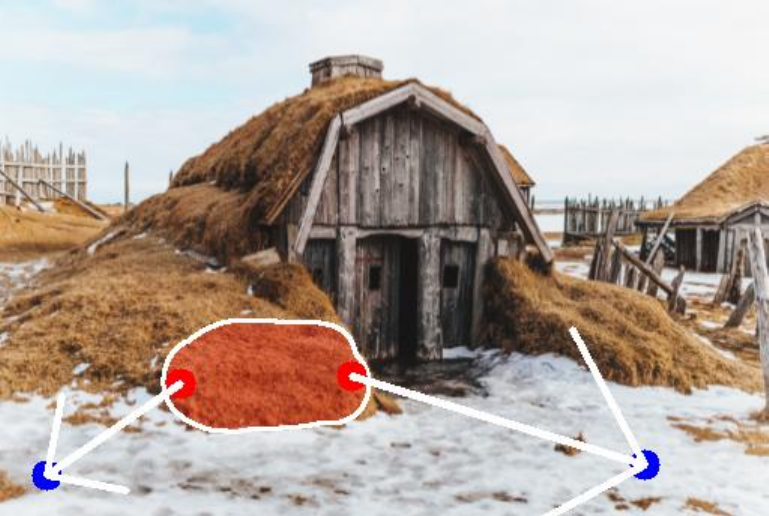} &
        \includegraphics[width=0.32\columnwidth]{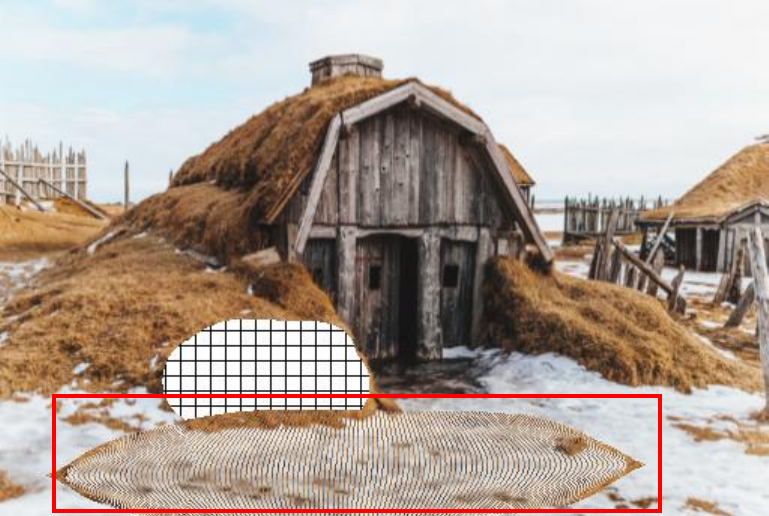} &
        \includegraphics[width=0.32\columnwidth]{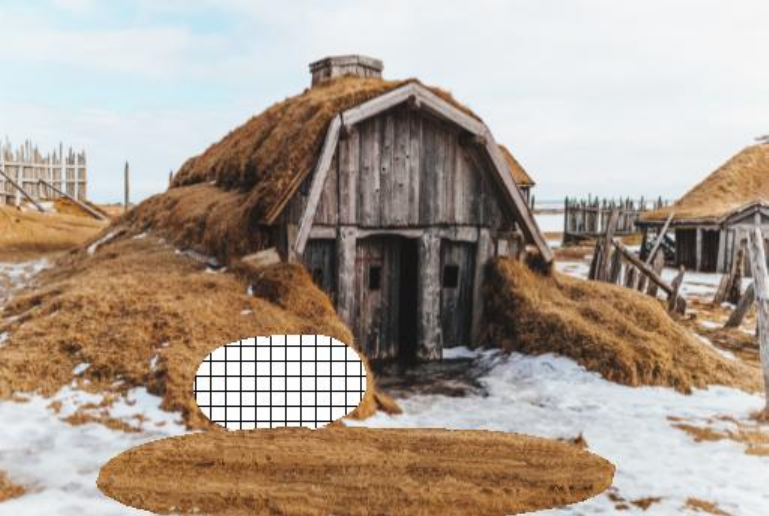} \\[-1ex]
        \includegraphics[width=0.32\columnwidth]{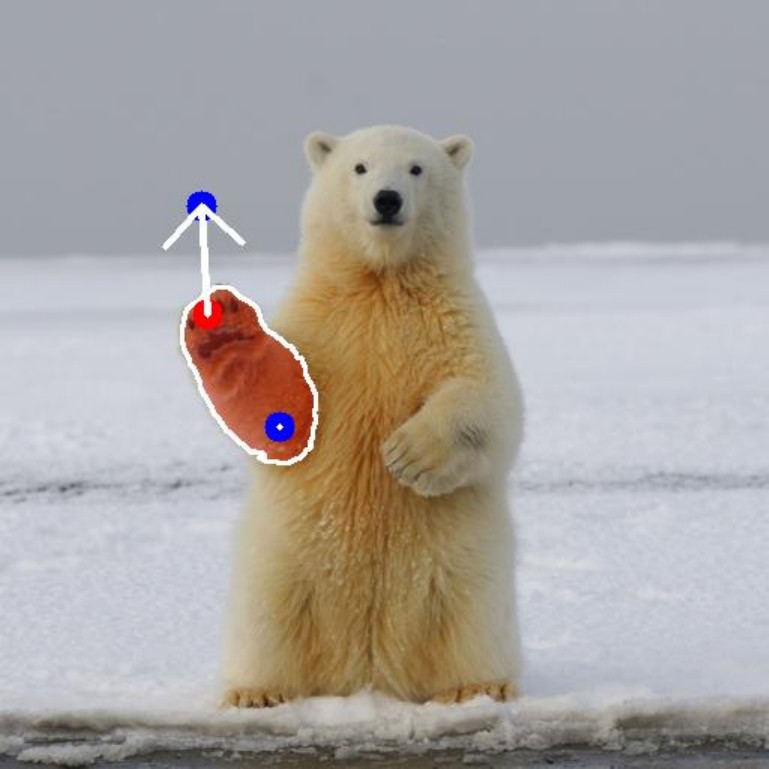} &
        \includegraphics[width=0.32\columnwidth]{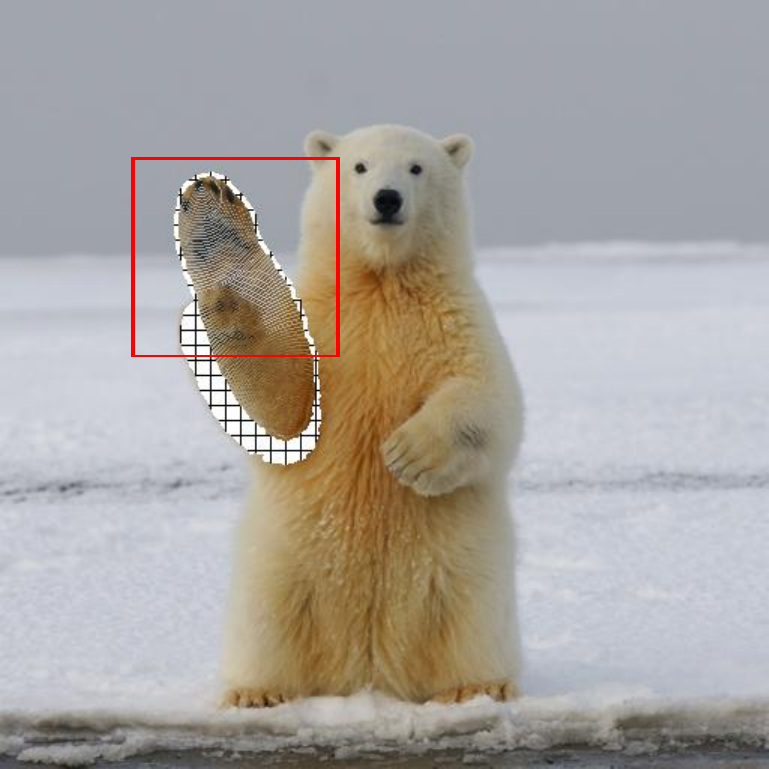} &
        \includegraphics[width=0.32\columnwidth]{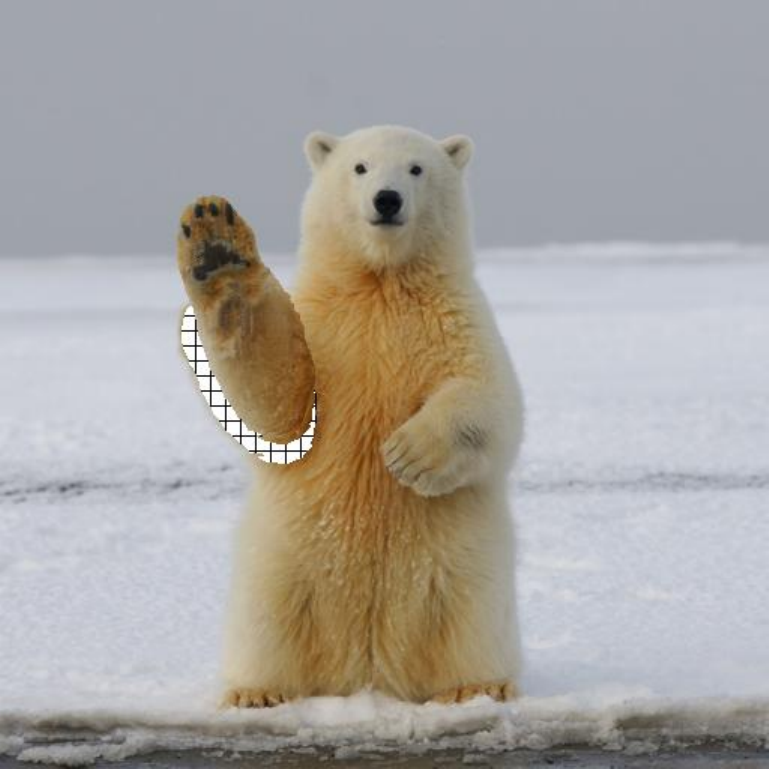}
    \end{tabular}
    \caption{Unidirectional \textit{vs.} bidirectional warping (ours). Our method fixes sampling gaps using backward pixel mapping.}
    \vspace{-2mm}
    \label{fig:bi}
\end{figure}

%% file: sec/5_conclusion.tex
\section{Conclusion}
In this paper, we introduce Inpaint4Drag, a novel approach that repurposes image inpainting for drag-based editing through pixel-space bidirectional warping. Unlike existing solutions that rely on general-purpose text-to-image models unoptimized for drag operations, our specialized separation of warping and inpainting effectively maintains pixel consistency while generating high-quality content in newly revealed areas. Our bidirectional warping algorithm and SAM-based boundary refinement provide real-time feedback for intuitive interaction. 
Experimental results show that Inpaint4Drag delivers superior performance while reducing processing time from minutes to milliseconds. Moreover, since Inpaint4Drag is compatible with any inpainting model, it can continuously improve alongside advancements in inpainting technology.

\clearpage
\noindent \textbf{Acknowledgments.} This work is supported by the Hong Kong Research Grants Council - General Research Fund (Grant No.: 17211024).

%% file: figs/chess.tex
\begin{figure*}[h]
\centering
\includegraphics[width=\textwidth]{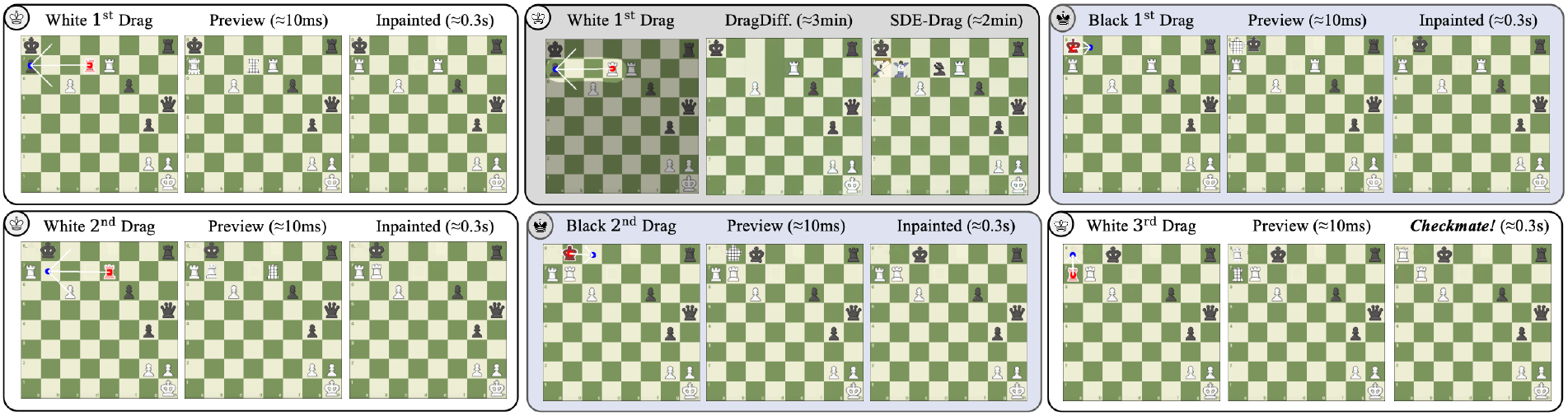}
\caption{Multi-round interactive drag editing demonstrated through a three-move checkmate sequence including five consecutive edits. Users select \textcolor{red}{deformable regions} (chess pieces) and drag them from \textcolor{red}{handle points} to \textcolor{blue}{target positions}. Grid overlays in the preview columns indicate areas requiring inpainting. Our method provides real-time preview ($\sim$10ms) of the warping effect, followed by high-quality inpainting results ($\sim$0.3s). Existing approaches typically require minutes for inference and fail during the initial interaction.}
\label{fig:chess_example}
\vspace{-4mm}
\end{figure*}